\documentclass[letterpaper]{article} 
\usepackage[draft]{aaai25}  

\usepackage{times}  
\usepackage{helvet}  
\usepackage{courier}  
\usepackage[hyphens]{url}  
\usepackage{graphicx} 
\usepackage{natbib}  
\usepackage{caption} 
\usepackage{algorithm}
\usepackage{algorithmic}
\usepackage{xcolor}
\usepackage{comment}
\usepackage{subcaption}

\newcommand{\elizabeth}[1]{\textcolor{blue}{[Elizabeth]: #1}}

\usepackage{todonotes}

\usepackage{newfloat}
\usepackage{listings}
\DeclareCaptionStyle{ruled}{labelfont=normalfont,labelsep=colon,strut=off} 
\floatstyle{ruled}
\newfloat{listing}{tb}{lst}{}
\floatname{listing}{Listing}
\title{Usage Governance Advisor: From Intent to AI Governance}

\author{
   Elizabeth M. Daly,
   Sean Rooney,
   Seshu Tirupathi,
   Luis Garces-Erice,
   Inge Vejsbjerg,
   Frank Bagehorn,
   Dhaval Salwala,
   Christopher Giblin,
   Mira L. Wolf-Bauwens,
   Ioana Giurgiu,
   Michael Hind,
   Peter Urbanetz
}

\affiliations{
    IBM Research\\

}

\usepackage{bibentry}

\begin{document}

\maketitle

\begin{abstract}

Evaluating the safety of AI Systems is a pressing concern for organizations deploying them. In addition to the societal damage done by the lack of fairness of those systems,
deployers are concerned about the legal repercussions and reputational damage incurred by the use of models that are unsafe. Safety covers both \emph{what a model does}; e.g.~can it be used to reveal personal information from its training set, and \emph{how a model was built}; e.g.~was it only trained on licensed data sets. Determining the safety of an AI system requires gathering information from a wide set of heterogeneous sources including safety benchmarks and technical documentation for the set of models used in that system. 
Responsible use is encouraged through mechanisms that advise the user in taking mitigating actions when safety risks are detected.
We present the \textit{Usage Governance Advisor} which identifies and prioritizes risks according to the intended use case, recommends appropriate models, benchmarks and risk assessments and most importantly proposes mitigation strategies and actions. 

\end{abstract}

%

\section{Introduction}
Organizations run a significant risk of legal, financial and reputational damage because of the misuse of AI systems when inadequately governed. AI governance is both an obligatory requirement and a strategic necessity.

Companies using AI in their products are duty-bound to implement responsible governance structures and have a strategic incentive to do so. Having a comprehensive understanding of one's AI systems mitigates threats posed by improper governance and ensures that operationally practices, e.g. monitoring, updating, are in line with evolving risks and regulations. The introduction of the EU AI Act~\cite{euaiact2024} and similar regulations, means that companies that implement responsible AI governance practices have a competitive advantage, e.g.~in their ability to quickly deploy \emph{safe} AI systems.

This paper presents the \textit{Usage Governance Advisor} a system supporting human-in-the-loop automation to ease the barrier of entry for governance in AI systems. Our solution prioritizes the use and deployment of Large Language Models (LLMs) as integral components of the governance and risk assessment framework. The broad and diverse capabilities of LLMs make it difficult to scope and monitor risks as even their creators may not be able to accurately predict their behaviour. The importance of risks depends on the model intended use. For example, if an LLM is being used in the context of entity extraction, it would be excessive to analyze the output of the LLM for the presence of toxic language. In the context of a text generation task, however, it is important to consider the toxicity of the language. Furthermore, if the AI system is used within a specific domain, e.g.~human resources decisions, then compliance with external policies might need to be included in the assessment. To address this we first elicit semi-structured information on the intended use of the foundation model. This allows
AI Tasks and associated risks to be identified. This information is then used to both recommend candidate models from an inventory and propose the appropriate risk assessment evaluations that should be applied. Finally, our solution uses the prioritized risk specification to evaluate the models across relevant dimensions, creating an audit trail of the intended use, the relevant risks identified, and the recommended mitigation actions and guardrails required before the solution should be approved for deployment. At the core of the solution is a risk taxonomy which is used to link the use to the model, the models to assessments, and the outcomes to mitigations. 
We employ a Knowledge Graph (KG) that organizes the risk-related information about AI systems within a dedicated ontology. We describe how we populate this KG and use it for decision making when assessing risks associated with an AI solution. We explain the development of the auto-assist questionnaire, model recommendation, risk evaluations and mitigation recommendations.

\begin{figure*}
    \centering
    \includegraphics[width=0.95\linewidth]{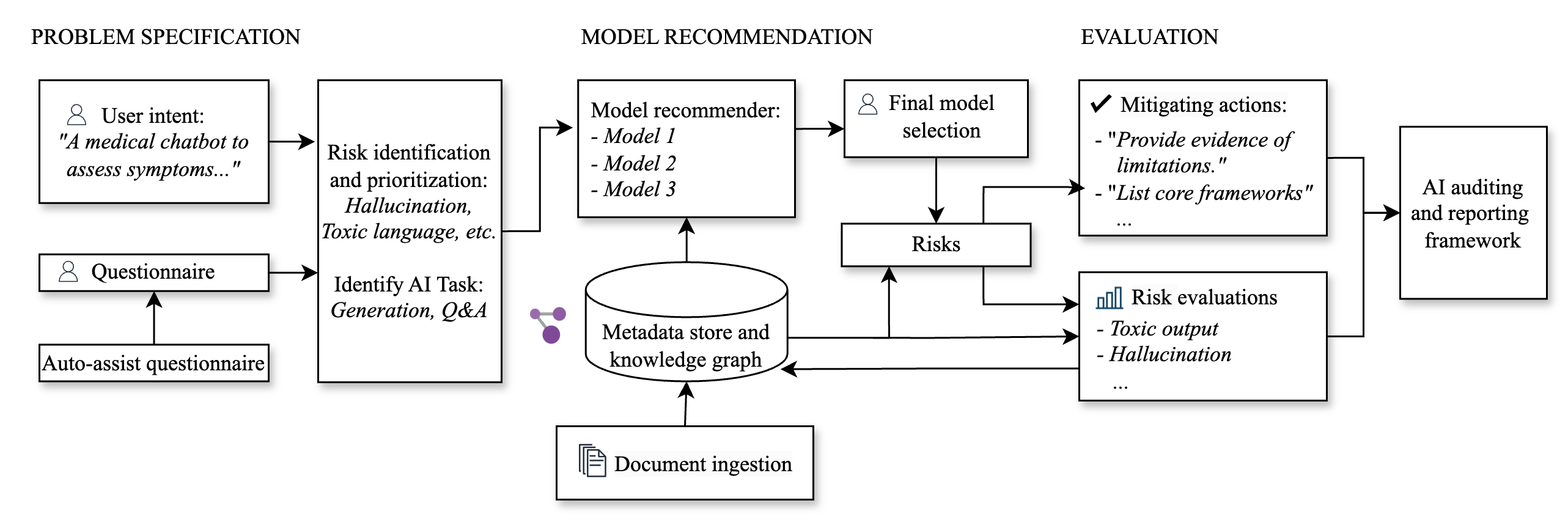}
    \caption{Usage Governance Advisor pipeline}
    \label{fig:arch}
\end{figure*}

\section{Related Work}

Implementing a comprehensive enterprise AI governance system is difficult due to the numerous stakeholders involved, e.g. Data scientists, Business owners, Risk Officers etc., and the need for human oversight to ensure trustworthiness in the decision-making process. Multiple studies have provided a conceptual overview of the disparate aspects of AI governance. For example,\cite{papagiannidis2023toward} describes the AI governance processes of three firms in the energy sector. The study describes the structural, relational, and procedural dimensions of governance by examining the knowledge gained through understanding the functioning of these companies. \cite{cihon2021corporate} discuss governance by considering the relevant stakeholders (e.g. managers, investors etc.) and their roles in AI governance. \cite{aangstrom2023getting} surveys the challenges companies face with AI implementation and governance. Finally, \cite{ferdaus2024towards} identifies the opportunities, challenges and limitations of trustworthy AI with a particular focus on LLMs. The paper also states the role of government initiatives for AI governance. 

KGs have been employed in legal governance by \cite{schwabe2020knowledge} to address concerns around trust, privacy, transparency, and accountability by incorporating legal resources into their framework. That paper describes a proof-of-concept framework 
that demonstrates how KG helps in addressing these concerns. The Artificial Intelligence Ontology (AIO)~\cite{joachimiak2024artificialintelligenceontologyllmassisted} defines AI concepts and relationships structured into six specific domains. Our ontology shares some common goals, however we have focused on enabling broad-based AI risk assessments. 

Despite the existence of conceptual frameworks that have explored individual facets of governing AI and LLMs, a comprehensive, systematic approach to analyzing and addressing risks has yet to be undertaken. Our work outlines a proof-of-concept system that describes what 
is required to achieve this goal.

\section{User scenario}

The Usage Governance Advisor (Figure \ref{fig:arch}) helps relevant stakeholders to uncover the risk profile of a use case and proposes appropriate models to perform the underlying AI tasks. In addition it recommends mitigating actions to offset the discovered risks. This information allows stakeholders to make an informed decision about the trade-offs involved in using a model. They can use the system to track risks for different models, and to meet their regulatory and compliance goals.
To illustrate how the system achieves these aims, consider a scenario in which AI developers are investigating the risks that are involved with the training and deployment of an AI system for a medical chatbot. The intent of this system is:

\emph{
``In a medical chatbot, create a triage system that assesses patients' symptoms and provides advice based on their medical history and current condition. The chatbot will identify potential medical issues, and offer recommendations to the patient or healthcare provider."}

The developers need to understand the potential risks associated with their use case and identify potential actions to mitigate those risks. The outputs of the system are designed to assist them in communicating this to other non-technical stakeholders in their organisation. The user inputs their use case to the system as an intent, and the system auto-fills a model usage compliance questionnaire using a Chain-of-Thought, LLM-as-a-judge approach to connect question/answer pairs to risks and identify an AI task. 

\begin{figure}
    \centering
    \includegraphics[width=0.95\linewidth]{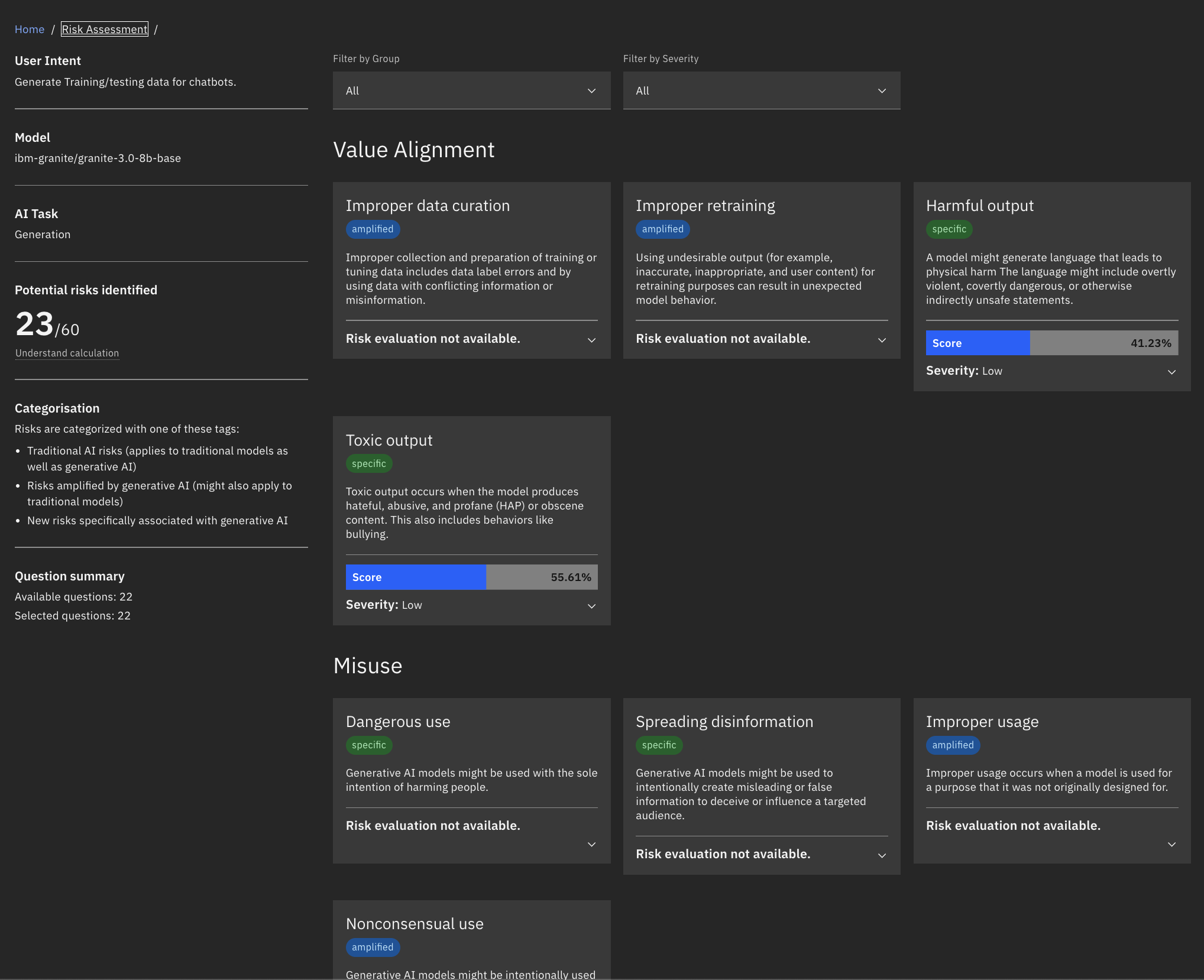}
    \caption{Risk Report}
    \label{fig:risk-report}
\end{figure}

\begin{figure}

        \centering
        \begin{subfigure}{.48\columnwidth}
            \includegraphics[width=\textwidth]{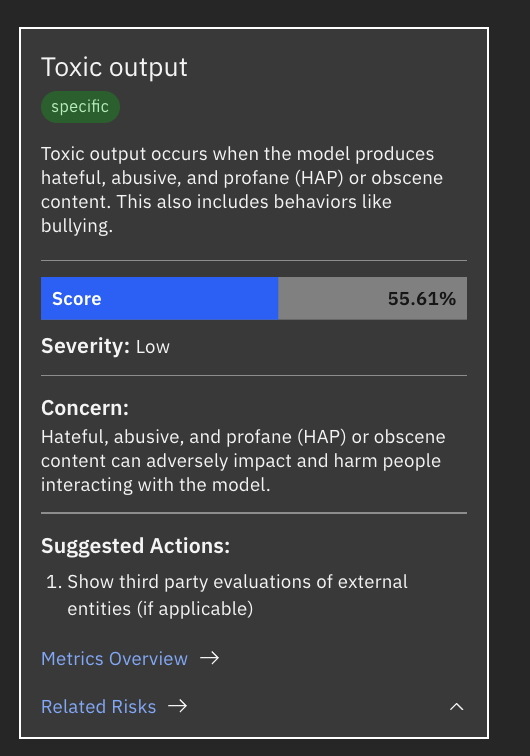}
            \caption[Measurable]{Measurable Risk}
        \end{subfigure} \hfill \begin{subfigure}{.48\columnwidth}
            \includegraphics[trim={0 130 0 0},clip,width=\textwidth]{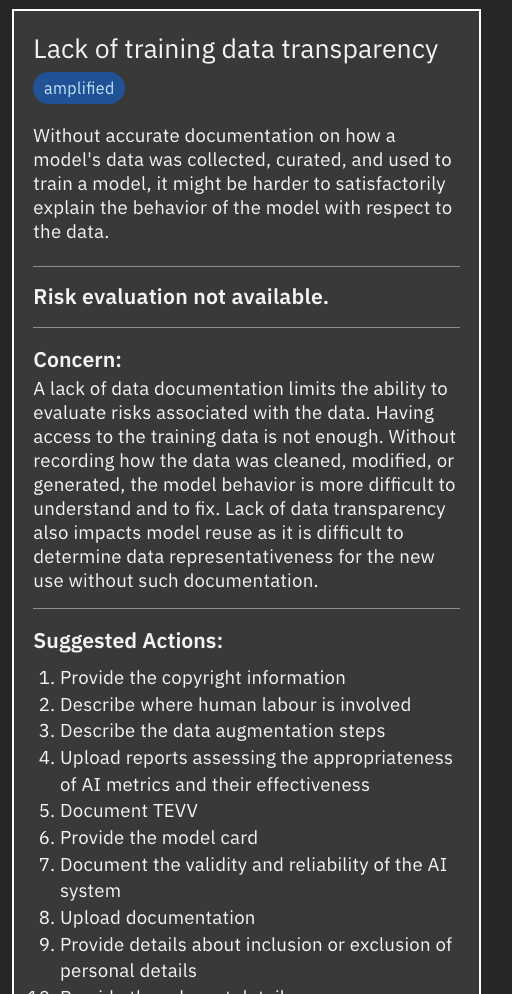}
            \caption[Non-Measurable]{Non-Measurable Risk}
        \end{subfigure}        
        \caption{Risk Cards}\label{fig:visuals}
        \label{fig:prototypical}
\end{figure}

The user inspects the questionnaire output (as displayed in Figure \ref{fig:risk-report}), and is asked to confirm the system's suggestions for the AI tasks (``Generation" in this case) or select alternate tasks. The list of potential risks, AI task, and intent are used to recommend different models in terms of their suitability, based on querying the knowledge graph for relevant benchmark evaluation scores. To provide additional transparency in the recommendation process, the evaluation scores are provided to the user as evidence of the reasoning behind the recommendation. The user confirms the suggested model is suitable for their case and the system computes a risk evaluation report. The user inspects the report to understand the categories of risks likely to be associated with their use case. For each risk identified, explanations are given about the concerns, allowing users to better interpret the system's recommendations. The user learns that ``Toxic output" is a risk associated with the output of the model within the use case. They can expand the card defining the benchmark to gather more details. Potential mitigating actions for that risk are proposed by the system in Figure \ref{fig:prototypical} and they can also see the scores of the related risk evaluations which have been run against the proposed AI model for the relevant risks, such as social bias and safety. The report is stored by the system and the developers can use it as a supporting tool to illustrate the risks to other non technical project stakeholders.


\section{System Overview}
The \emph{Usage Governance Advisor} provides workflows taking the user-intent and multiple sources of information to 1) create semi-structured information through compliance questionnaires, 2) prioritize these risks based on the inferred answers and 3) recommend appropriate models. The system collects risk evaluations associated with the proposed models in relation to the intended use case. The system recommends potential mitigation strategies and action items, leveraging a knowledge graph to help organize the information about AI models.

\subsection{AI Systems Knowledge Graph (KG)} 
We use a KG to structure disparate information about AI model usage risk. The details of the defined ontology, ingestion process and entity/relationship extraction are described in the following sections.

\subsubsection{Ontology Definition}
Underlying the system is a common ontology organizing concepts and relations within the domain of AI, with a focus on AI Governance concepts such as lineage, data sets, licensing, technical characteristics, evaluation results, and risk assessment. Given the rapid evolution of the field of AI,  there is a need for unifying concepts, mappings and the ability to integrate existing vocabularies. 


\begin{figure}
    \centering
    \includegraphics[width=\linewidth]{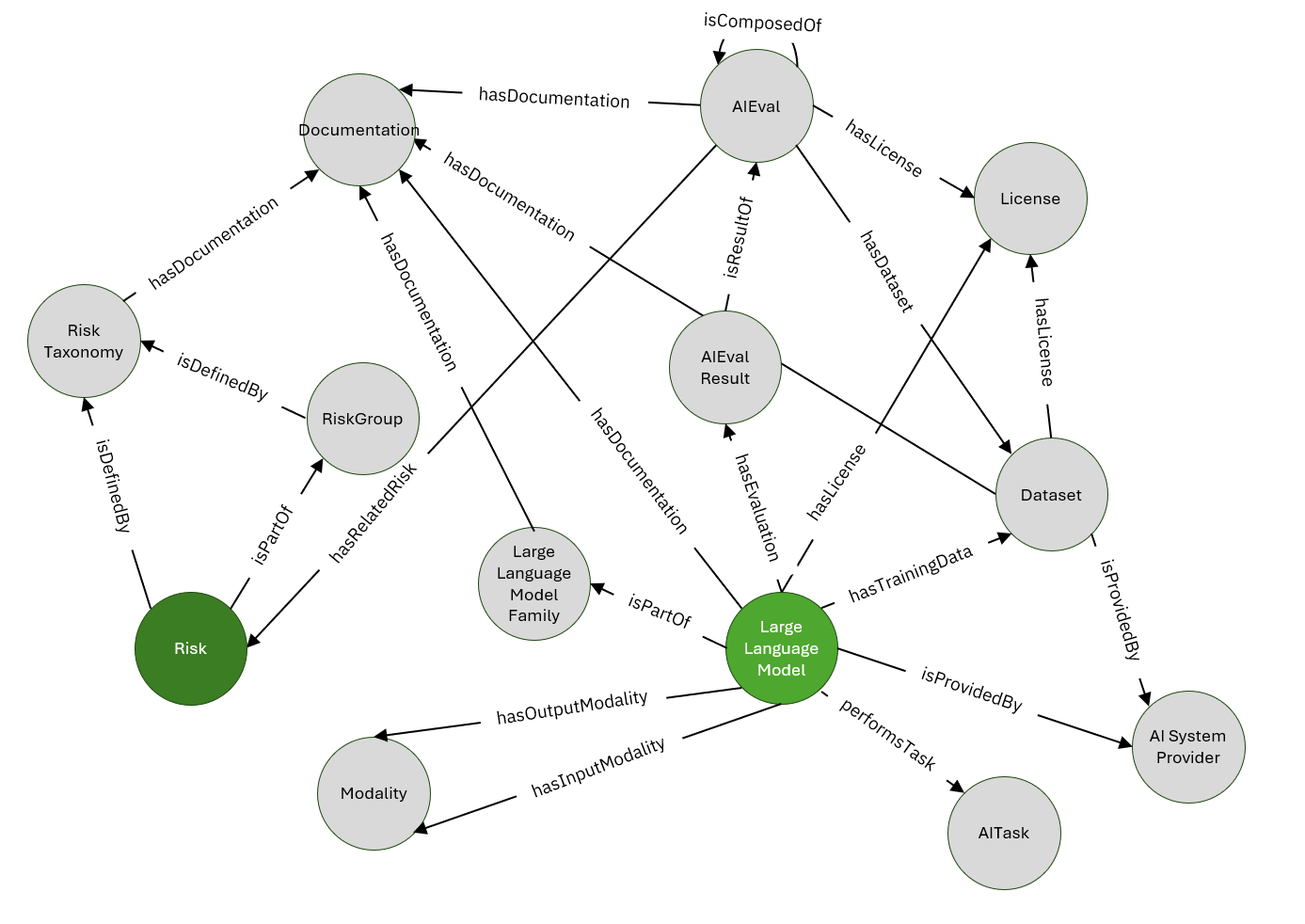}
    \caption[Schema]{AI Governance Ontology}
    \label{fig:schema}
\end{figure}

The AI ontology is defined with the LinkML modeling language and framework~\cite{moxon2021linked}. LinkML uses a commonly used definition language, i.e.~YAML, enabling it as a simple lingua-franca among an inter-disciplinary AI governance teams.  The framework provides generators for converting LinkML models into other schema languages such as JSON Schema, SQL data definitions and RDF allowing it to act as bridge between ontologies and the broad ecosystem of technologies found in modern development and run-time environments. Key elements and relationships in the ontology are shown in Figure \ref{fig:schema}.
The EU AI Act emphasizes AI systems rather than individual models. Consequently, the class \emph{AiSystem} is comprised of one or more \emph{AiModels}. \emph{LargeLanguageModel}, a sub-class of \emph{AiModel}, is trained on \emph{Datasets}. Both models and datasets have \emph{Licenses}. \emph{AiEvaluations} are associated with  \emph{AiEvalResults} and \emph{Risks} which can belong to a \emph{RiskTaxonomy}. \emph{Risks} may have mappings to \emph{Risks} in other taxonomies. 

We materialize this ontology in the KG by ingesting information from multiple sources. Essential information about the AI models includes: \emph{how the model was built}, e.g.~technical documents associated with the model and \emph{how the model behaves}, e.g.~output from dedicated test frameworks.

The KG encompasses both a domain graph of AI entities and relationships as well as an evidence graph associating AI entities with evidence (e.g.~AI evaluation results). The relationships between these two graphs allows the user to find the source of a given piece of information in the KG. This provides confidence in the entity/relationships extracted by the ingestion process. These are generated as subject-verb-object triples.

\subsubsection{Applying the Ontology to Risk Mapping}

The Usage Governance Advisor supports a risk assessment of an AI system. To model AI risks, an existing AI risk ontology, AIRO~\cite{airo}, is used for the base risk vocabulary. The IBM AI Risk Atlas \cite{riskatlas} was chosen for the base risk taxonomy as it is used widely in industry. However, communities (e.g.~OWASP AI security) and regulatory jurisdictions (e.g.~US Federal Systems) may require different risk taxonomies. As a result we defined mappings between the  base taxonomy and three leading risk taxonomies: the NIST Gen AI Profile~\cite{nist}, the MIT AI Risk Database~\cite{airiskrepo} and OWASP~\cite{owasp}.

\begin{figure}
    \centering
    \includegraphics[width=0.95\linewidth]{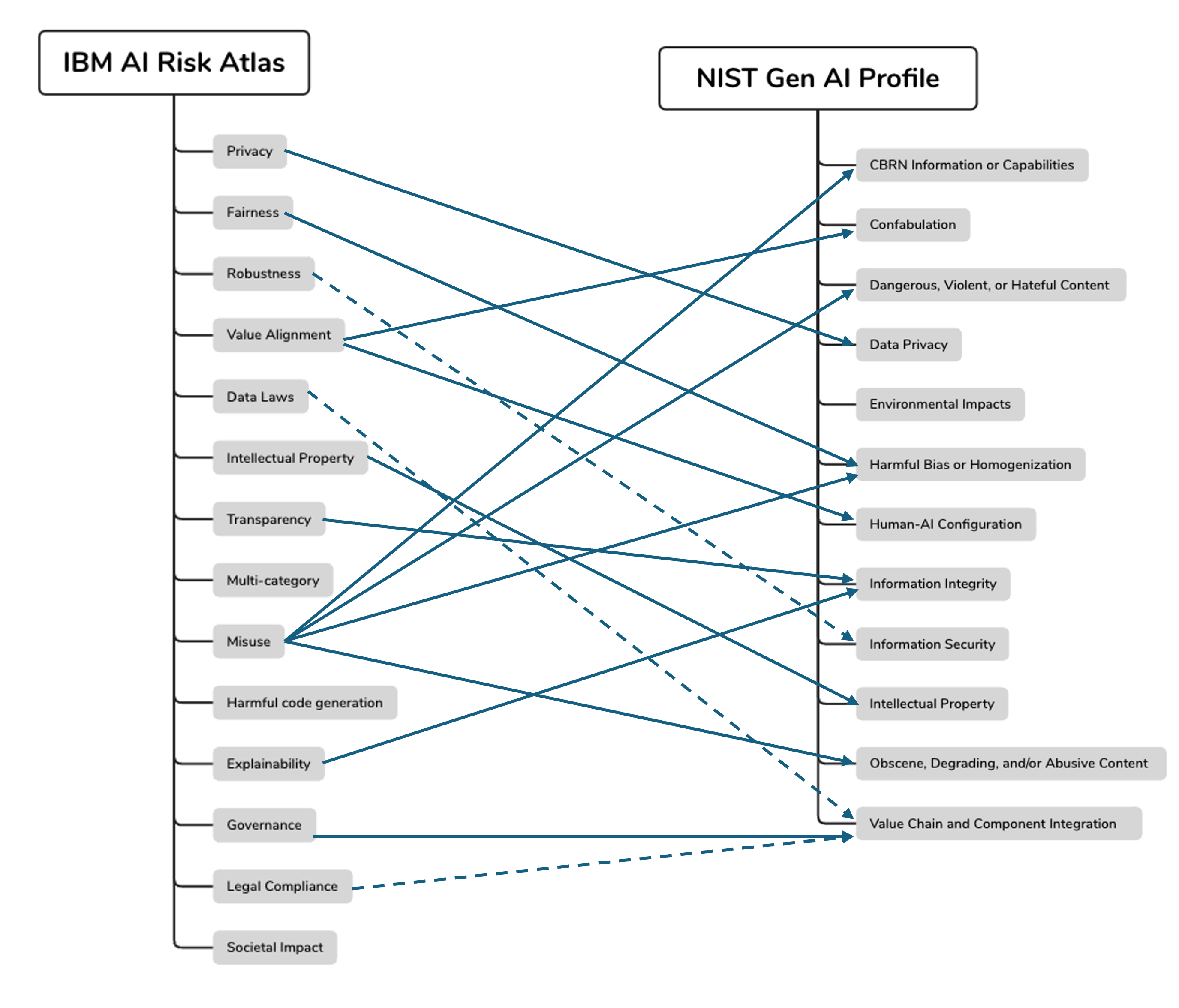}
    \caption{Risk Taxonomy Mappings }
    \label{fig:risk-map}
\end{figure}

It is difficult to draw simple isomorphic equivalences between these taxonomies due to the differences in their focus, granularity and structure. For example, Figure \ref{fig:risk-map} shows mappings between the IBM AI Risk Atlas and the NIST Gen AI Profile. We use the Simple Knowledge Organization System (SKOS)~\cite{skos} schema to capture the different relationships of \texttt{skos:closeMatch}, \texttt{skos:exactMatch}, \texttt{skos:broadMatch} and  \texttt{skos:relatedMatch}.

\subsubsection{KG Construction}
\label{sec:kg}

As much of the information stored in the KG exists in unstructured documentation, our system uses a generative AI pipeline to ingest data from these sources. This allows
the system to scale as it doesn't require the manual population of the KG by a human, but involves some probability of incorrect information being added. We will consider this issue further in the analysis section.

We maintain both a domain graph containing the extracted facts and an evidence graph relating the sources of information for those facts and an indication of our confidence in their truth. This pipeline consists of: ingestion, entity/relationship extraction, and augmenting with evidence and confidence measures. The entire pipeline is coordinated using the langchain~\cite{langchain} framework.

\textbf{Ingestion:} We first classify documents to determine their nature. This helps in identifying entities and relationships for extraction, i.e.~if the document is about a specific AI Model e.g.~LLAMA-3, then most relationships will link that entity to other entities, e.g.~its license. Simple term frequency highlights the document’s main subject such as an AI model or regulation.  This focuses the extraction process and narrows the relevant subset of the ontology to consider. Similar work is discussed in \cite{PeaTMOSS}.

After classifying the document, we decompose the text of the document into chunks that can be treated in isolation by the generative pipeline. The process retains the relationship between chunks of the document. The chunks are labeled with additional information such as the classification, section title, and their location in the document. Some overlap with the previous and next chunk is included. We adapt the method described in~\cite{Mishra_2024} for chunking PDFs in a manner suitable for Retrieval-Augmented Generation (RAG)~\cite{NEURIPS2020_6b493230} systems. 

\textbf{Entity/Relationship Extraction:} To organize the information from the documents into the KG, it needs to be converted to entities and relationships following the types described in the ontology. Pragmatically, to reduce the complexity of the problem, we make the following assumptions: 1) the entity types relevant to the document have been extracted in the ingestion phase, so we do not need to consider the entire ontology when processing a document 2) there is no need to extract the specific relationship types between two entities, only as to whether two entities are related; the relationship should be implicit from the ontology. This reduces the expressiveness of the ontology but we believe that this isn't prohibitive for the domain being modelled.

A further simplification is that we do not try to identify entities with properties as defined in the ontology, but rather we promote properties to entities themselves. We found that
it is simpler to perform only simple entity recognition rather
than identifying entities \emph{and} their associated properties together. So for example, the energy usage of a AI model
is a distinct entity, rather than a property of that model.
The reason for this is that when relevant information
is available in different chunks spread across a large document it is simpler to relate distinct entities found
in different chunks rather than updating properties of the same entity.

\subsection{Auto-assist questionnaire}
\begin{figure}[t]
    \centering
    \includegraphics[width=0.85\linewidth]{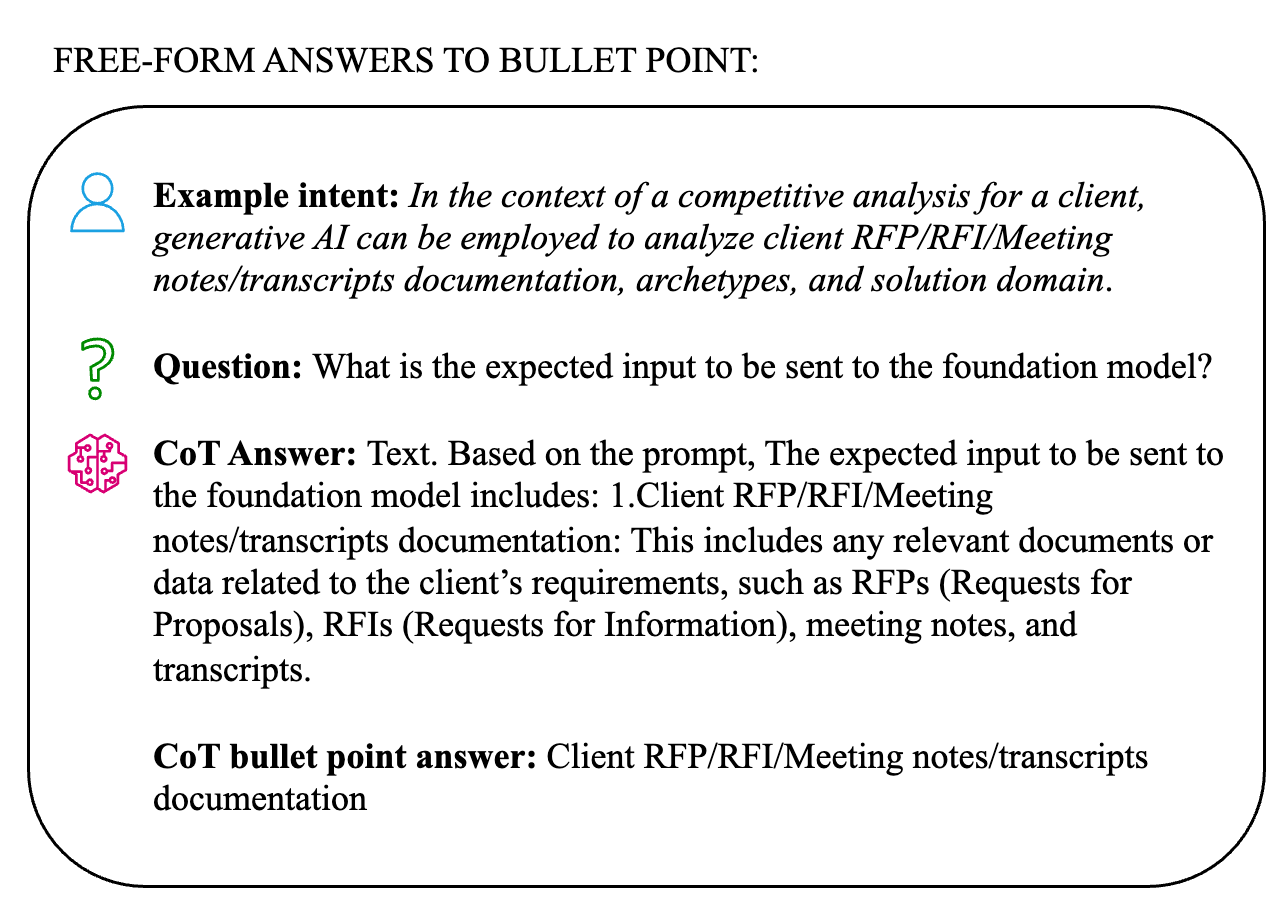}
    \caption{Intent and conversion of free-form answers to bullet point answers from the questionnaire.}
    \label{fig:sample_intent}
\end{figure}

\begin{figure}
    \centering
    \includegraphics[width=0.85\linewidth]{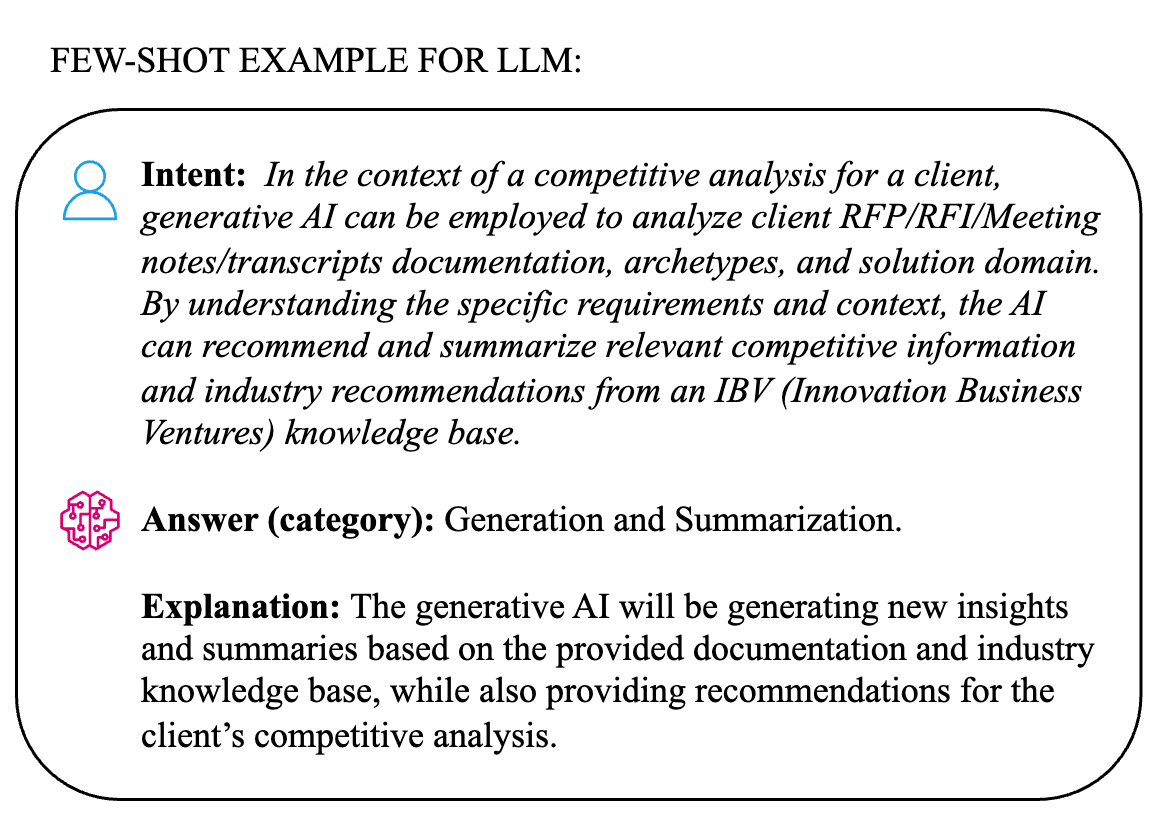}
    \caption{One of the few-shot examples provided to the LLM for the question of identifying the category based on intent.}
    \label{fig:few_shot}
\end{figure}

To help manage the deployment of LLMs, organizations follow a process to identify and mitigate risks. A key component of this process is the completion of questionnaires to aid in assessing the risks associated with the specific AI use case and model being deployed.
This process can be time-consuming and cumbersome for end-users, who must navigate lengthy questionnaires prior to  deployment. To address this challenge, we have developed an auto-assist functionality that utilizes a Chain-of-Thought (CoT) approach and few-shot examples to assist users in completing questionnaires as shown in Figure \ref{fig:sample_intent} and \ref{fig:few_shot}. This functionality provides suggestions for answers to compliance questions based on the user's initial intent. 
The auto-assist functionality supports three types of questions: multiple choice, binary and freeform.

\subsection{Risk Prioritization}
\begin{figure}
    \centering
    \includegraphics[width=0.85\linewidth]{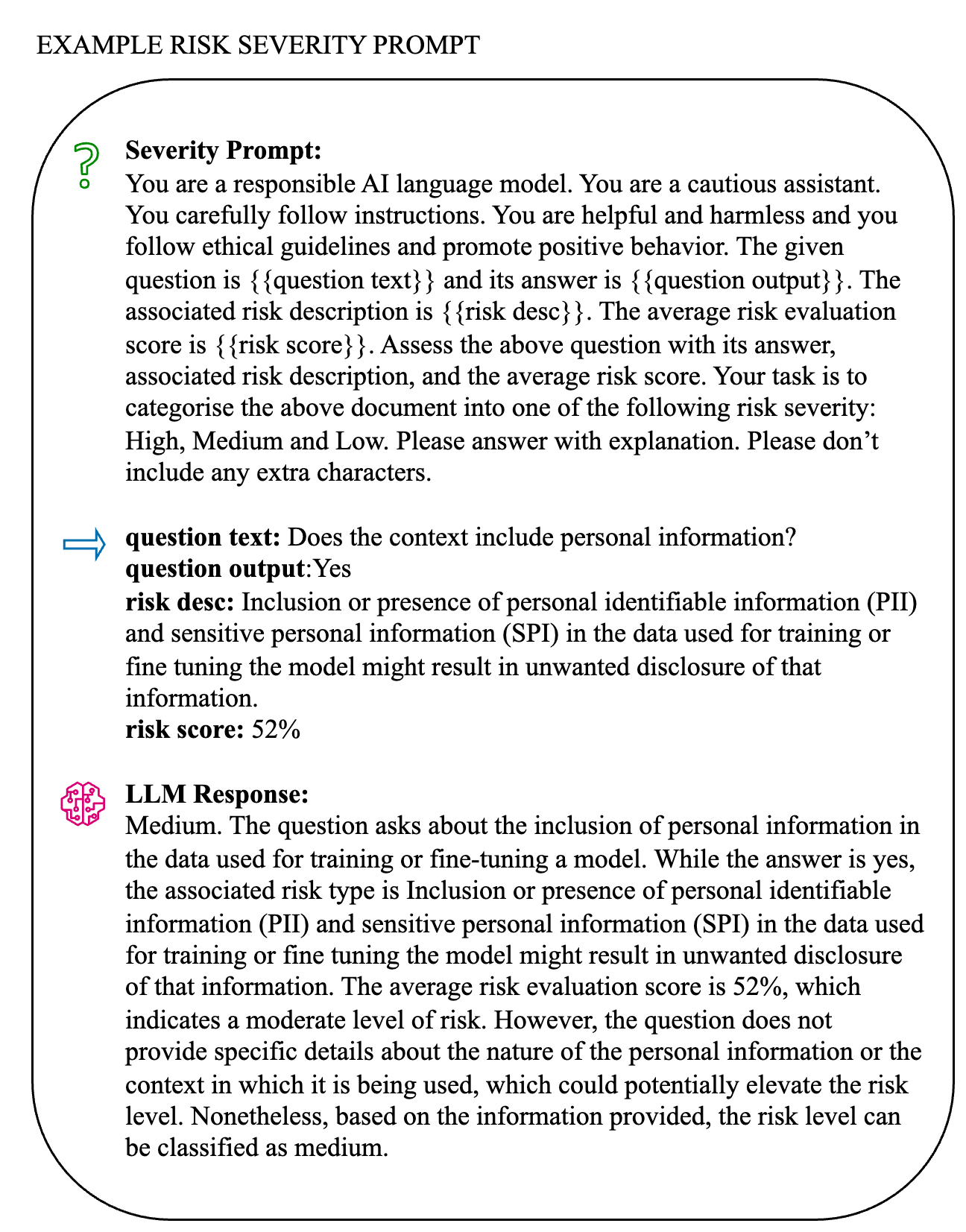}
    \caption{Risk severity prompt with an LLM response.}
    \label{fig:severity_prompt}
\end{figure}

Risks are identified by analyzing the answers to risk
specific questionnaires. 
We use an LLM-as-a-judge approach to connect questions/answer pairs to risks and whether specific answers reduce or amplify a risk \cite{zheng2023judging}. Each risk is assigned a severity level, which classifies the  potential impact of a specific risk into three classes: High, Medium, and Low. 
Figure \ref{fig:severity_prompt} presents an example prompt used to classify the risk severity level based on the risk description, the questions/answer pairs from which the risk was inferred, and the normalized average score of all the individual scores associated with the risk. In this example, the context variables in the prompt are substituted with the relevant factors discussed above. When assessing the risk, the LLM evaluates whether the response to the question helps mitigate the risk in relation to the question's context and the risk description. 

\subsection{Model Recommendation}
The model recommender selects models from a model inventory for use in a specific use case with an appropriate risk profile.

Customer policy is used to define acceptance criteria for recommended models. The KG also contains historical risk evaluations, i.e., benchmarks and questionnaires, that quantify risk exposure of AI models. This can be combined with information about a deployed model's behavior from logging data.


The recommender combines both the prioritized risk, evaluation results and acceptance criteria to estimate a  total risk value for each candidate model, identifying the lowest risk model for the use case. If a candidate model has already been specified for a similar use case, the model recommender can compare this model with other challenger models and list strengths and weaknesses of the candidate model. In case of incomplete information about risk evaluations, additional automated risk evaluations are proposed.

\subsection{Automated Risk Evaluations}
General purpose LLMs have a broad range of capabilities and with this comes a wide variety of risks~\cite{weidinger2021ethical,schillaci2024llm}. The research community has responded by creating an ever-increasing number of benchmarks \cite{zhang2023safetybench,xu2023sc,zhu2024promptbench}. Given the rapidly evolving collection of benchmarks and evaluations we developed a flexible approach with two goals. First, given the large number of risk dimensions it is untenable to run all evaluations for all use cases As a result, we run only a subset of prioritized risk evaluations based on the prioritized risks identified in the risk identification process. Second, we allow easy creation of new assessments leveraging the Unitxt framework \cite{bandel2024unitxt}. Usage Governance Advisor maintains linkages between risk atlas definitions and Unitxt specified risk evaluations. The catalogue is regenerated frequently to include new benchmarks required for the risk assessment evaluation. 



Individual risks may be specified as quantitative or categorical. A quantitative risk is a numeric score from a safety benchmark.
As these scores are diverse and are not easily meaningfully combined we normalize the raw score into three numbers: 
 \emph{1, 0, -1}, where \emph{1=Above average}, \emph{0=Average}, \emph{-1=Below average}. The average being calculated from applying the same measure to a collection of reference models. The customer policy is then used to compute a weighted average 
 to create a single number. The policy enables risk stakeholders to prioritize
 different risks for a specific use-case.

Categorical risks are mapped to the same scale: \emph{1, 0, -1}.
Where \emph{1=Known and acceptable}, \emph{0=Not known}, \emph{-1=Known and unacceptable}. The term \emph{acceptable} is configurable via the deployers' policy. This is similar to the approach used in the Stanford Transparency Index~\cite{bommasani2023foundationmodeltransparencyindex} that measures whether some aspect of a model is known.  For example, the customer policy might define that the license is known and corresponds to a set of approved licenses, or that the model vendor claims that approval for usage has been obtained for all data sets. Categorical risks are used to filter the models that are considered.

\subsection{Mitigation Recommender}
Identified risks are used to recommend mitigation strategies. Our solution recommends two types of mitigation strategies: deployment guardrails and manual mitigating actions. 

\subsubsection{Guardrails}
Detectors and guardrails are common practice for LLMs in deployment \cite{inan2023llama,achintalwar2024detectors,magooda2023framework}. Detectors can act as real-time filters to mitigate harmful outcomes (e.g. generating toxic output). In a similar manner for evaluations, Usage Governance Advisor maintains links between risk dimensions and guardrails. Assessments conducted as part of the automatic risk evaluation can be augmented to include a post-processing guardrail step to measure the impact of a given guardrail on mitigating a given risk. As a result, the system can recommend mitigation guardrails that should be used for any system deployment along with evidence of the anticipated impact of such a guardrail. 

\subsubsection{Manual Curation of Recommended actions}
The different types of risks identified require different types of actions. 
Whereas some actions are directly identifiable from the risk, other actions require different types of (human) action. An example of the former is the following: \emph{Detected situation: For all data used in building the model, copyright status is not disclosed. Risk: Data usage rights (Data Provenance), Action: Provide copyright information).} 
An example for the latter case: \emph{Detected situation: A model’s limitations are not demonstrated. Risk: Explaining output (Output bias). Action:1.)  Contact developer/model provider to demonstrate the model’s limitations. 2.) Understand whether limitations impact the output in expect use context. 3.) Implement guardrails. 4.) Assess model output after guardrail implementation. 5.) If 4 not satisfactory, reconsider use for specific use context.}
These examples also illustrate the importance of manual curation of recommendation actions: depending on the risk identified, the suggested actions can be a straightforward description of how to obtain the missing information. In other cases, a series of actions with different dependencies has to be followed. Often, these actions require human intervention including intervention in existing business processes. Thus, as mitigation actions can be very context-specific, it is also advisable to test and assess the mitigation actions that Usage Governance Advisor provides, within the specific context of use. 

A risk can only be marked as resolved once the deployment guardrail has been executed and/or the mitigating action implemented, with the actions taken documented and assessed. This documentation also allows for evaluating the effectiveness of these measures over the model's lifetime.


\section{Analysis} 
We evaluate two foundational workflows used in the Usage Governance Advisor that themselves use generative AI.

\subsection{Knowledge Graph Construction}
The KG is populated by extracting entities/relationship in the ontology from technical documentation. 
The only way to scale this approach is through the use of highly generic fully automated ingestion processes.
We evaluate different approaches using generative AI against manually obtained ground truth from the \emph{granite-8b-code-base-4k} AI model card in Hugging Face~\cite{granite-8b-code-base-4k}. 
The generated KG is then measured against the ground truth by considering
entity/relationship triples and calculating how many of the ground truth triples
exist in the prediction (recall) and how many of the predicted triples exist in the ground truth (precision). 
Note that precision can be reduced by both incorrect triples, i.e hallucinations, as well as correct but irrelevant ones. 

We provide the F1 metric relating the two for completeness. This is the same approach used in~\cite{10.1007/978-3-031-47243-5_14}. 

We assume that for two triples to match, the two related entity types must match, the entity labels (names) must also match and the relationship is \emph{implicit} from the ontology; i.e., it does not matter what the relationship type is, only that the two entities in the triple are related. For example, if an \textit{AI Model }entity is related to a \textit{License} entity; then independently of whether the extraction process labeled the relationship is \textit{has}, \textit{released-under}, \textit{uses }etc. we assume that the triple refers to the license of said AI model.

Results are provided in Table~\ref{experiments} for a selection of different extraction methods. One can see that running the process multiple time increases the F1 score due to the random nature of the LLM entity identification, i.e. additional entities are correctly identified in subsequent runs and incrementally added to the graph. The process was run up to 15 times, however, performance reached the upper limit within 2-4 passes, which are reported in the table.

Preliminary results showed that having an exact match for triples in the ground truth and triples generated by an LLM is too strict a metric. As an example, consider the following: A triple \texttt{['ibm research', 'organization', 'granite-8b-code-base-4k', 'aimodel’]} does not match \texttt{['granite-8b-code-base-4k', 'aimodel', 'ibm research', 'organization’]}, although only the direction of the relationship is distinct. 
To support these cases, we extend the pipeline with an LLM-as-a-judge solution using \emph{granite-3-8b-instruct} which improved results as shown in Table~\ref{experiments}. The judge improve recall and precision by identifying matches that a simple lexical comparison would miss.

\emph{Granite-3-8b-instruct} as Judge recovered the most information, and was correct around 80\% of the time.
Note that other than the ontology itself the extraction process is independent of the nature of the entities and relationships, allowing it be reused even as the ontology evolves. In order to test an upper-bound on the accuracy of the extraction process we also used a set of highly specific prompts, labelled as \emph{Overfitted}. These do better than all the generic approaches but only by about 8\% than the best motivating the usability of the generic approach. 

The use of generative AI currently does not allow 100\% certainty about the validity of the generated information. In our view there should be
a measure of certainty given about extracted facts with suitable links to evidence that a human could check and verify if the certainity is below an acceptable threshold. Measuring the accuracy of model utterances is an active area of AI research.

\begin{table}[]
\begin{scriptsize}
\begin{center}

\begin{tabular}{|l|l|l|l|l|l|}
\hline
\emph{Entity/Relationship Extraction} & \emph{Match} & \emph{Runs} & \emph{P} & \emph{R} & \emph{F1} \\ 
\hline\hline
LLMGraphTransformer                & Exact & 1 & 0.25 & 0.36 & 0.29 \\
\cline{2-6}
with mixtral-8x7b-instruct-v01     & Exact & 3 & 0.26 & 0.52 & 0.35 \\
\cline{2-6}
                                   & Judge & 4 & 0.70 & 0.76 & 0.73 \\
\hline\hline
GraphRag with                      & Exact & 1 & 0.27 & 0.4 & 0.32 \\
\cline{2-6}
mixtral-8x7b-instruct-v01          & Exact & 2 & 0.30 & 0.6 & 0.40 \\
\cline{2-6}
                                   & Judge & 4 & 0.71 & 0.76 & 0.73 \\
\hline\hline
RAG with                           & Exact & 1 & 0.25 & 0.16 & 0.19 \\
\cline{2-6}
mixtral-8x7b-instruct-v01          & Judge & 1 & 0.75 & 0.6  & 0.66 \\
\hline\hline
Custom prompt with                 & Exact & 1 & 0.17 & 0.48 & 0.25 \\
\cline{2-6}
mixtral-8x7b-instruct-v01          & Judge & 1 & 0.77 & 0.88 & 0.82 \\
\hline\hline
Custom prompt with                 & Exact & 1 & 0.20 & 0.64 & 0.31 \\
\cline{2-6}
granite-3-8b-instruct              & Judge & 1 & 0.76 & 0.84 & 0.80 \\
\hline\hline
Overfitted prompt                  & Exact & 1 & 0.61 & 0.64 & 0.62 \\
\cline{2-6}
                                   & Judge & 1 & 0.92 & 0.92 & 0.92 \\
\hline
\end{tabular}
\caption{Precision (P), Recall (R) and F1 metric for the triples obtained with different methods.}
\label{experiments}
\end{center}
\end{scriptsize}
\end{table}

\subsection{Auto-assist questionnaire}
We evaluated the effectiveness of the Chain-of-Thought (CoT) approach in auto-assisting questionnaire completion with synthetically generated user intents. We generated 42 artificial user intents and corresponding answers using an LLM. These answers were validated by human-annotated ground truth. 
 Accuracy is used to measure binary and dropdown questions. To evaluate free-form questions, the answers generated by CoT were condensed to bullet points to better support accuracy measures and compared with the ground truth. 
Figure \ref{fig:sample_intent} provides an illustrative example of user intent, 
response, and bullet-point summary generated by the CoT approach and consequent user summarization. 
Table \ref{table:example_questions} shows a subset of the questions used to assess the effectiveness of CoT approach. 
For each question, we provided few-shot or Chain-of-Thought reasoning examples to facilitate consistent response generation (see Figure \ref{fig:few_shot}).

\begin{table}[]
\resizebox{\columnwidth}{!}{%
\def\arraystretch{1.1}%
\begin{tabular}{|l|}
\hline
\textbf{Example questions}                                                                                                                                         \\ 
\hline
\begin{tabular}[c]{@{}l@{}}What category of use does your use request fall under? \\ Classification, Recognition, Generation, Summarization, Ideation, Question/Answer, \\ Search and Information Seeking, other.\end{tabular} \\ \hline
\begin{tabular}[c]{@{}l@{}}What is the expected input to be sent to the foundation model?\end{tabular}                     \\ \hline
\begin{tabular}[c]{@{}l@{}}Does the context include personal information? \end{tabular}                                             \\ \hline
\end{tabular}%
}
\caption{Example questions from the questionnaire.}
\label{table:example_questions}
\end{table}

\begin{table}[t]
\centering
\resizebox{\columnwidth}{!}{
\begin{tabular}{|l|l|l|l|}
\hline
\textbf{Question type} & \textbf{Zero-shot} & \textbf{Few-shot/CoT} & \textbf{Few-shot/CoT} \\
                       &                    &       \textbf{(1 choice)}  & \textbf{(user choice)} \\ \hline
    Dropdown questions &  0.40 & 0.73 & 0.915 \\ \hline
    Binary questions & 0.6 & 0.81 & -  \\ \hline
    Freeform questions & 0.73 & 0.81 & 0.905  \\ \hline
\end{tabular}
}
\caption{Accuracy for the different category of questions.}
\label{table:accuracy}
\end{table}

The performance of the CoT approach was evaluated using  \emph{granite-3-8b-instruct}. The results presented in Table \ref{table:accuracy} 
demonstrate that the CoT method outperforms zero-shot inference for all three types of questions, yielding significantly higher accuracy. 
Furthermore, when users are offered the option to select from a subset of suggested answers, the performance of CoT is enhanced. 
Notably, the upper limit for additional suggestions provided by the LLM has been set to two options above the ground truth. If the LLM proposes more than this threshold, it is considered an incorrect response.



\section{Perspectives and Future Work}
Effective AI governance requires a multifaceted approach that considers diverse stakeholder perspectives (service providers, users, auditors) to mitigate risks. Governance rules are inherently dynamic and subject to frequent updates. In addition, new models are continuously emerging in the market. This makes it challenging to ensure that all risks have been appropriately considered. This paper explores the potential of knowledge graphs as a solution for governing AI systems. By integrating new rules, models, and mitigation strategies seamlessly, while maintaining relationships with existing information, KGs can provide a comprehensive framework for managing AI governance complexities.
We present an essential first step in addressing regulatory and compliance requirements for deploying AI models into production, supporting alignment with evolving laws and regulations. Real-time monitoring and auditing of AI system performance are also crucial to identify risks associated with AI failures or misuse. To further enhance the framework, future work will focus on developing a robust mechanism for continuously auditing and verifying AI system performance through integration of real-time monitoring capabilities. Additionally, we aim to incorporate policy guidelines from stakeholders directly into the knowledge base, enabling a more tailored approach to AI governance. By addressing these aspects, the framework can provide a comprehensive and proactive approach to AI governance, ensuring responsible deployment and continuous improvement of AI systems.
 

\bibliography{aaai25}

\end{document}